\pdfoutput=1
\documentclass[11pt]{article}
\usepackage[final]{acl}

\usepackage[utf8]{inputenc}   
\usepackage[T1]{fontenc}    	
\usepackage{url}              		 
\usepackage{booktabs}       	
\usepackage{amsfonts}       	
\usepackage{nicefrac}       	  
\usepackage[final,nopatch=footnote]{microtype}      	

\usepackage{graphicx}
\usepackage{color}
\usepackage{rotating}
\usepackage{tabularx}
\usepackage{pdflscape}
\usepackage{amsmath,amsthm,amssymb}
\usepackage{algorithm,algorithmic}
\usepackage{bbm,dsfont}
\usepackage[caption=false]{subfig}
\usepackage{wrapfig}
\usepackage{cancel}
\usepackage{enumerate,cases}
\usepackage{thmtools,thm-restate}
\usepackage{mathtools}

\usepackage[none]{hyphenat}
\usepackage{multirow}
\usepackage{makecell}
\usepackage{setspace}
\usepackage{pifont}
\usepackage{array}
\usepackage{enumitem}
\usepackage{bm}
\usepackage{lineno}     

\usepackage{inconsolata}
\usepackage{times}
\usepackage{latexsym}

\allowdisplaybreaks

\usepackage{silence}
\usepackage{hyperref}		 
\hypersetup{
}
\usepackage[capitalize]{cleveref}

\usepackage{todonotes}
 
\newcommand{\Train}{\mathcal{T}}
\newcommand{\Inference}{\mathcal{I}}

\newcommand{\F}[1]{\boldsymbol{F}\!\left( #1 \right)}
\newcommand{\T}[1]{\boldsymbol{T}\!\left( #1 \right)}
\DeclareMathOperator*{\LLM}{LLM}

\renewcommand{\phi}{\varphi}
\renewcommand{\epsilon}{\varepsilon}

\newcommand{\el}{\end{flushleft}}
\newcommand{\bl}{\begin{flushleft}}

\theoremstyle{plain}

\theoremstyle{definition}

\theoremstyle{remark}

\title{
    Uncovering Scaling Laws for Large Language Models via Inverse Problems
}

\author{
    \textbf{Arun Verma\textsuperscript{1}},
    \textbf{Zhaoxuan Wu\textsuperscript{1}},
    \textbf{Zijian Zhou\textsuperscript{1,2}},
    \textbf{Xiaoqiang Lin\textsuperscript{2}}, \\
    \textbf{Zhiliang Chen\textsuperscript{2,3}},
    \textbf{Rachael Hwee Ling Sim\textsuperscript{2}}, 
    \textbf{Rui Qiao\textsuperscript{1,2}},
    \textbf{Jingtan Wang\textsuperscript{2,3}},
    \textbf{Nhung Bui\textsuperscript{2}}, \\
    \textbf{Xinyuan Niu\textsuperscript{2,3}}, 
    \textbf{Wenyang Hu\textsuperscript{2,5}}, 
    \textbf{Gregory Kang Ruey Lau\textsuperscript{2,6}},
    \textbf{Zi-Yu Khoo\textsuperscript{2,7}},
    \textbf{Zitong Zhao\textsuperscript{2}},  \\
    \textbf{Xinyi Xu\textsuperscript{2,3}},
    \textbf{Apivich Hemachandra\textsuperscript{2}},
    \textbf{See-Kiong Ng\textsuperscript{4}},
    \textbf{Bryan Kian Hsiang Low\textsuperscript{1,2}}
    \\~\\
    {\small
        \textsuperscript{1}Singapore-MIT Alliance for Research and Technology  \qquad
        \textsuperscript{2}Dept. of Computer Science, National University of Singapore
    }
    ~\\{\small
        \textsuperscript{3}Agency for Science, Technology and Research \qquad
        \textsuperscript{4}Institute of Data Science, National University of Singapore
    }
    ~\\{\small
         \textsuperscript{5}SAP \qquad \textsuperscript{6}CNRS@CREATE \qquad \textsuperscript{7}AI Singapore \qquad \textbf{Correspondence:} \href{mailto:lowkh@comp.nus.edu.sg}{lowkh@comp.nus.edu.sg}
    }
}

\begin{document}
    \maketitle

    \begin{abstract}
        Large Language Models (LLMs) are large-scale pretrained models that have achieved remarkable success across diverse domains. These successes have been driven by unprecedented complexity and scale in both data and computations. However, due to the high costs of training such models, brute-force trial-and-error approaches to improve LLMs are not feasible. Inspired by the success of inverse problems in uncovering fundamental scientific laws, this position paper advocates that inverse problems can also efficiently uncover scaling laws that guide the building of LLMs to achieve the desirable performance with significantly better cost-effectiveness.
    \end{abstract}

    \section{Introduction}
    \label{introduction}

LLMs represent a paradigm shift in artificial intelligence, embodied by their unprecedented levels of complexity and scale in both data and computations, and their demonstrated generalization capabilities across a wide array of tasks and domains, such as natural language processing, computer vision, coding, gaming, among many others~\citep{Bommasani2021FoundationModels,claude3, ArXiv23_openai2023gpt, nijkamp2023codegen2,dubey2024llama,reid2024gemini}.
These remarkable successes result from the amalgamation of several \emph{input} ingredients, including high-quality and diverse training data, advanced modeling techniques, skillfully designed training procedures, and effective inference schemes~\citep{Wei2022_CoT,Anthropic-system-prompts,apple-system-prompt}. 
The intricate interactions among these ingredients are not fully understood, yet they collectively influence the overall performance of large language models.
To advance the development of high-performance and cost-effective models further, it is essential to uncover the underlying scaling laws that govern these \emph{interactions}.
More importantly, designing an LLM that achieves desirable performance \emph{under resource constraint} is an inherently complex challenge, as it requires the careful selection and combination of data, model architecture, training procedures, and inference strategies.

As an example, when building an LLM specifically for GSM8K~\citep{arXiv21_cobbe2021training} (i.e., grade school math benchmark), several design principles must be considered: 
(i) The training data should contain ample examples that foster language understanding and reasoning capabilities to ensure that the LLM can learn the nuances of math problems presented in natural language;
(ii) the model architecture should be complex enough to process sequential inputs (since each question in GSM8K is described in natural language) and generate the required output formats, such as multiple-choice questions or detailed natural language explanations;
(iii) the training procedure should be designed to allow the model to effectively acquire task-specific knowledge from the data (e.g., suitably defined loss functions tailored for solving math problems); and
(iv) the inference scheme should guide the LLM toward generating accurate and desired outputs, as demonstrated by techniques like Chain of Thought (CoT)~\citep{Wei2022_CoT}, ReAct~\citep{ICLR23_yao2023react}, and Tree of Thoughts~\citep{NeurIPS23_yao2023tree}. 
\begin{figure*}[!ht]
    \centering
    \includegraphics[width=0.67\linewidth]{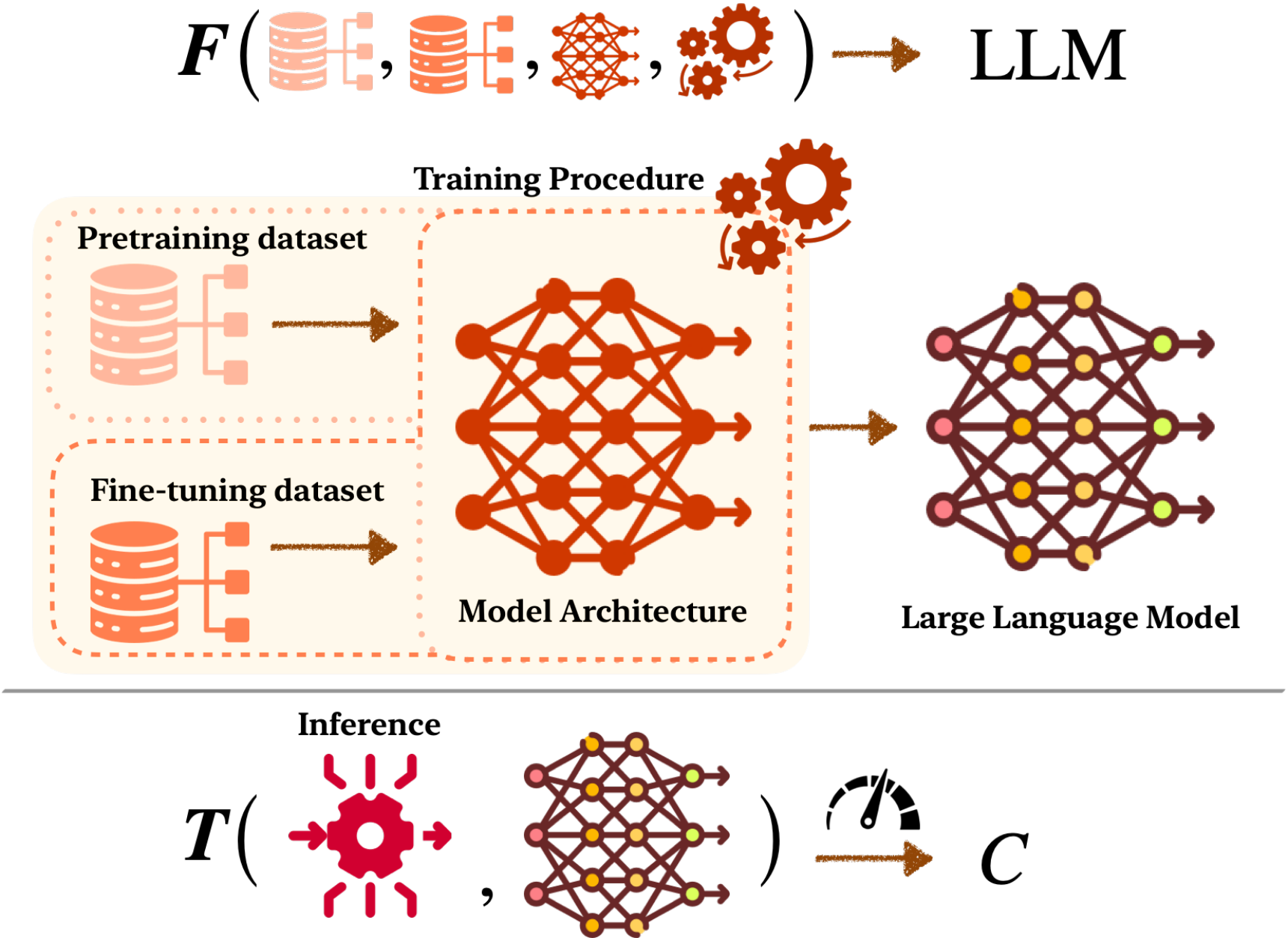}
    \vspace{-1mm}
    \caption{
        Forward processes in large language models.
        The forward process generates an LLM from key input ingredients and components: datasets, model architecture, and the training procedure. During inference time, other ingredients such as the prompt examples would affect the desired performance metric $C$.
    }
    \label{fig:diagram_forward}
    \vspace{-3mm}
\end{figure*}

Due to the scale of the required data and modern model architectures, creating an LLM instance is an extremely costly process, e.g., GPT-$4$ costs over \$$100$ million~\citep{GPT4_100million} while the cost for Gemini Ultra is estimated at over \$$191$ million~\citep{Gemini_cost_HAI2024}.
This high expense makes building better LLMs through brute-force trial and error prohibitively costly. 
In contrast, DeepSeek V3 achieved state-of-the-art performance with just \$$5.6$ million by optimizing training protocols and architecture~\citep{arXiv24_liu2024deepseek,DeepSeek2025}.
Thus, it becomes necessary to \textit{uncover underlying scaling laws} (e.g., the required composition and minimum size of training data or model architecture) that help \textit{build LLMs with the desired performance and significantly better cost-effectiveness}.

To this end, \textbf{\em we advocate examining the class of inverse problems for LLMs}. 
Inverse problems involve determining unknown parameters of an underlying model from observational data, a concept crucial in scientific and engineering domains~\citep{groetsch1993inverse, vogel2002computational, chadan2012inverse, gazzola2018forward}. 
Tackling the inverse problems is a tried-and-true methodology for inferring and uncovering fundamental scientific laws from observations. 
For example, Kepler's laws of planetary motion were derived from the observed motion of Mars and its elliptical orbit; Newton's law of universal gravitation was based on the empirical effects of Kepler's laws; and in modern quantum mechanics, Schr\"odinger's wave equation was inferred from electron diffraction experiments.
Inspired by these successes, inverse problems offer a powerful approach for uncovering the underlying scaling laws behind the behavior of LLMs.

A typical approach to tackle an inverse problem involves using a \textbf{\em forward process} to obtain observation data given some specified input and latent parameter values. 
However, this forward process is often costly. 
The \textbf{\em inverse problem}, which involves identifying the latent parameter values that are consistent with a given set of observation data, is inherently very challenging due to the complexity of the search space and the lack of solution uniqueness. 
In the LLM context, the inverse problem requires finding the optimal combination of input ingredients (i.e., data, model architecture, training procedures, and inference schemes) to build the LLMs that achieve desirable performance, while forward processes refer to the costly training of LLMs and running model inference for task execution and evaluation.

Formally, let $\Train$ denote the training ingredients, such as the dataset, model architecture, and training procedure, and let $\Inference$ represent the ingredients of the inference scheme (e.g., prompting method). 
Note that $\Train$ includes \emph{both} pretraining and fine-tuning, and it affects the LLM's model parameters, whereas $\Inference$ typically does not alter these parameters. 
Let $\F{\Train} \rightarrow \LLM $ denote the process of creating an LLM by executing the computation following the specified ingredients $\Train$ (i.e., forward process).
Let $\T{\LLM, \Inference} \rightarrow C$ represent the evaluation of the LLM on a task using the inference scheme $\Inference$, resulting in a performance metric $C$. 
Therefore, we have the following two forward processes:
\vspace{-1.5mm}
\begin{subequations}\label{eq:forward}
    \begin{align}
        &\F{\Train} \rightarrow \LLM\ , \label{eq:F} \\
        &\T{\F{\Train}, \Inference} \rightarrow C\ . \label{eq:T}
    \end{align}
\end{subequations}
\vspace{-6mm}

\noindent These two forward processes are illustrated in \cref{fig:diagram_forward}. 
To understand how these forward processes function, consider the above example of building an LLM for the GSM8K task, which assesses various design principles related to data (i.e., pretraining and fine-tuning datasets), model architecture, and training procedure. 
These ingredients are included within $\Train$ and used in the creation process of an LLM as $\F{\Train} \rightarrow \LLM$. 
Subsequently, the trained LLM, along with the inference ingredients $\Inference$, is evaluated on the GSM8K task as $\T{\LLM, \Inference} \rightarrow C$. 
Here, $\T{\cdot}$ includes both the evaluation metric (e.g., accuracy) and the evaluation dataset (i.e., GSM8K questions), and $C$ is thus representing the accuracy of the trained LLM on the GSM8K benchmark.

Given practical constraints such as limited data and computational resources, tackling the inverse problems to uncover end-to-end scaling laws may be overly ambitious.
Therefore, as a first step, we consider simplified inverse problems by fixing certain ingredients or focusing on a manageable subset of the problem space.
Specifically, this position paper frames the following classes of inverse problems in the context of LLMs:
\vspace{-3mm}
\begin{itemize}
	\setlength\itemsep{-0.25em}
    \item In~\cref{sec:data_selection}, we frame \textbf{Data Selection} as an inverse problem, focusing on integrating multiple data modalities, exploiting commonly used yet non-differentiable metrics, and enhancing selection efficiency. Solving this problem is expected to improve downstream performance while reducing the need for extensive human feedback.

    \item In~\cref{sec:inference}, we frame \textbf{Inference Optimization} as an inverse problem and focus on the inference scheme used in conjunction with trained models.
    Solving this problem ensures trained models are adapted to underlying downstream tasks using minimal resources, without needing to modify their parameters.

    \item In~\cref{sec:unlearning}, we frame \textbf{Machine Unlearning} (MU) verification and MU for LLMs to achieve desired performance metrics as inverse problems. 
    Solving these problems ensures data owners that their deletion requests are fulfilled and assures model owners that harmful data are removed.
\end{itemize}

    \section{Data Selection}
    \label{sec:data_selection}

The recent successes of LLMs have been driven by training on massive and heterogeneous datasets~\citep{Xu2024DataCentric}.
For example, LLaMA 3 was trained on 15 trillion multilingual tokens~\citep{dubey2024llama}. 
Previous works have established scaling laws that link data quantity to model performance~\cite{kaplan2020scaling,NeurIPS22_hoffmann2022training,zhai2022scaling,wu2024inferencescalinglawsempirical,chen2025duetoptimizingtrainingdata}.
However, more recent studies~\citep{xia2024less, wang2024helpful_freeshap,chen2025duetoptimizingtrainingdata,qiao2025grouprobust,wang2025nice} demonstrate that strategically selecting data subsets can improve the performance of both LLMs and multi-modal LLMs (MLLMs) in a way even surpassing the conventional scaling laws, particularly in domains like computer vision~\citep{sorscher2022beyond}.
This naturally raises key questions: How does model performance scale with data quantity when data selection methods are used for MLLMs? Furthermore, how do the scaling laws vary across different stages of MLLM training, such as pretraining, fine-tuning, and alignment?

We formulate data selection as an inverse problem of $\T{\F{\Train}, \Inference} \rightarrow C$. 
The goal is to understand how the quantity of selected training data (in $\Train$) scales with the desired MLLM performance ($C$).
For example, we might want to identify the minimal dataset required to train an MLLM to achieve specific performance metrics under optimal data selection. 
Therefore, efficient data selection can significantly reduce computational costs by prioritizing informative and representative data, thereby improving training efficiency without sacrificing performance. 
Furthermore, these scaling laws should be general enough so that they are applicable to a family of data selection methods instead of specific implementations (e.g., the family of influence functions~\citep{koh2017understanding} versus its implementation DataInf~\citep{kwon2024datainf}).

\subsection{Data Selection for Multi-Model LLMs}
\begin{figure}[!ht]
    \centering
    \includegraphics[width=\linewidth]{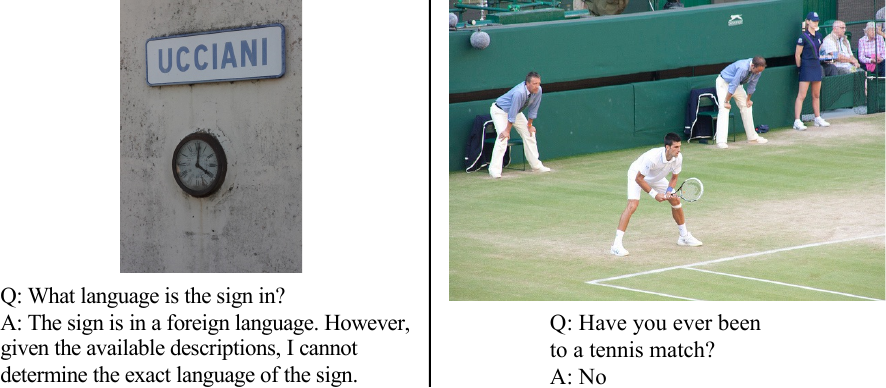}
    \caption{
        Examples of low-quality samples include instances where the question-answer pair fails to capture the key elements of the image or has limited relevance.
    }
    \label{fig:low_quality_samples}
    \vspace{-3mm}
\end{figure}

The remarkable successes of LLMs have led to the development of MLLMs that integrate advanced visual processing capabilities~\citep{dai2023instructblip,liu2023llava, zhu2023minigpt4}. 
However, the rapid growth of the MLLMs and their multi-modal nature have led to instruction-tuning datasets that often rely on automated or template-based content, resulting in relatively poor-quality and redundant datasets~\citep{liu2024more}, as illustrated in \cref{fig:low_quality_samples}.
To address this challenge, introducing smaller yet high-quality datasets can potentially maintain or even improve the performance of MLLMs. 
Traditional data pruning methods often require repeated gradients retrieval~\cite{park2023trak} or extensive memory for storage~\cite{yang2023dataset}, both of which become impractical for MLLMs due to their massive model sizes and data volumes. Conventional attribution methods, such as semivalues~\citep{ghorbani2019datashapley,Zhou2023pasf}, the influence function~\cite{koh2017understanding, kwon2024datainf}, and TracIn~\cite{pruthi2020estimating} have not been widely adapted for MLLMs.
This naturally raises a question: How to perform effective data selection for MLLMs while considering both image and text features?

Previous efforts have approached the problem as a large-scale data selection challenge, focusing on external evaluators such as established criteria~\cite{wei2023instructiongpt4} or intrinsic features~\cite{chen2024visionlanguage,liu2024more}. 
For example, \citet{xia2024less} demonstrated using a small subset of textual training data can achieve the same performance as the full dataset.
The next step is to propose relatively more compute-friendly methods and generalize them to the large-scale domain of MLLMs, improving upon the standard power law scaling.
The core objective of data selection research is to identify the techniques that enable training to scale efficiently and effectively with increasing amounts of data~\citep{TMLR24_albalaksurvey}.

In addition, some training data points may rely primarily on a single modality (e.g., cases where images alone suffice to answer the questions).
Would the scaling laws of data selection differ across different modalities, and would any particular modality have a stronger impact on the performance? 
To address these inquiries, one can potentially employ feature attribution methods like Integrated Gradients~\cite{sundararajan2017axiomaticattributiondeepnetworks} to attribute the score of each training data point to specific modalities.
The multi-modal nature of data introduces an additional layer of complexity, rendering the adaptation more challenging than its conventional application in computer vision tasks. 
Analyzing these modality-specific scores will help better understand the relative importance of each modality and how these modalities influence the overall performance, ultimately uncovering a universal scaling law for all modalities.

\subsection{Data Selection for LLM Fine-tuning with Non-differentiable Performance Metrics}
Commonly used data selection methods in LLMs are often the gradient-based data attribution methods~\citep{han-etal-2020-explaining, schioppa2022scaling,yeh2022first,grosse2023studying, wang2024data,zhou2024datavalue}, such as influence functions~\citep{kwon2024datainf} and TracIn~\cite{xia2024less}, which quantify the impact of each data point on model parameters and next-token prediction loss. 
However, non-differentiable metrics $C$, such as semantic similarity with the ground truth~\citep{cer2017semeval}, BLEU score~\citep{papineni2002bleu, sellam2020bleurt}, reward models~\citep{ouyang2022training}, and LLM-as-a-judge~\citep{zheng2024judging}, are commonly used to evaluate the LLM performance in practice. 
This discrepancy between the metrics used for data selection and the metrics employed for evaluating the LLM performance can result in sub-optimal performance.
Therefore, we advocate for research on how to select data for LLM fine-tuning when optimizing for commonly used but non-differentiable evaluation metrics.
This problem is non-trivial because, unlike influence functions, there is no straightforward way to compute the effect or gradient of the non-differentiable evaluation metric with respect to the model parameters and training data.

One promising approach is the integration of non-differentiable evaluation metrics into the data selection method using reinforcement learning techniques, for instance, the policy gradients from the REINFORCE algorithm~\citep{williams1992simple,wang2025nice}.
By serving as a surrogate for ``gradients'' of the non-differentiable evaluation metrics with respect to model parameters, these methods can lead to a novel data selection method that directly optimizes desired (non-differentiable) evaluation criteria, thereby directly uncovering the underlying scaling laws that link the amount of training data to model performance.

\subsection{Data Selection for LLM Alignment}
Existing works have shown that LLM responses often do not immediately align with user intent after pretraining or fine-tuning, as LLMs can generate untruthful, unuseful, and even harmful contents~\citep{bai2022training}.
However, recent successes~\citep{nisan2020learning,ouyang2022training} in training LLMs using human feedback has improved alignment between user intent and LLM responses (i.e., achieving the desired alignment performance metric $C$) via methods like  Reinforcement Learning with Human Feedback (RLHF)~\citep{ouyang2022training} and Direct Preference Optimization (DPO)~\citep{rafailov2024direct}. 
Achieving the desired alignment depends heavily on obtaining high-quality human feedback (i.e., human labeling), which is \textit{costly} and requires a large amount of feedback to ensure effective alignment training (i.e., RLHF/DPO). 
This challenge has motivated the development of a heuristic-based approach~\citep{muldrew2024active} that aimed at efficiently selecting a subset of LLM responses for human feedback. 
However, this heuristic-based approach lacks a principled foundation, leading to the following question: How to actively select the LLM responses for human feedback in a principled way to minimize the amount of feedback required while ensuring effective RLHF/DPO alignment training?

To address this problem, one can consider designing theoretically grounded acquisition functions specifically tailored for efficient LLM alignment~\citep{ICLR25_verma2025neural,arxiv25_verma2025active}. 
Such acquisition functions should explicitly account for variations in pretrained data and model architecture, which can lead to potentially different preferences for responses depending on these factors. 
Specifically, the acquisition functions need to incorporate the DPO process and quantify the uncertainty for the difference between the latent scores of two prompt-response pairs, where the latent scoring function is defined using the LLM itself~\citep{rafailov2024direct,lin2025activedpo}. 
Uncovering scaling laws to efficiently acquire high-quality and diverse training data from LLM users can significantly reduce the budget required for data collection.

\subsection{Joint Optimization for Data Selection} 
Previous discussions focus on data selection for a single training stage. However, different training stages improve different aspects of the model capability, and combining them can further improve the performance~\cite{ke2023continual}. Specifically, continued pretraining can be used to keep the knowledge of the model updated~\cite{ke2023continual,jindal2024balancing} while instruction fine-tuning can improve its ability to follow natural language instructions~\cite{wei2022finetuned}. 
Thus, a question naturally arises:
How to decide the ratio of data points used in different stages under a fixed number of data points? 
A joint optimization approach can be plausible to find the optimal ratio~\citep{jindal2024balancing}. 
Finding this optimal ratio helps uncover the underlying scaling laws of optimal data selection across different training stages, changing the scaling law of model performance $C$ with respect to the dataset size.

Recent results from training LLMs for low-resource languages such as SEA-LION~\cite{sea_lion_2024} demonstrate that combining continued pretraining with instruction fine-tuning achieves superior performance.
On the other hand, selecting the best training data also depends on the LLM/MLLM architecture. Existing model selection works~\cite{raschka2018model_model_selection, wang2021rethinking_model_selection,xia2024less} typically seek to find the optimal model architecture given fixed training data or the other way around. Therefore, producing the best-performing LLM/MLLM requires us to jointly select the most appropriate data and model architecture. Hence, an important research direction will be to develop algorithms that jointly select data and model architecture \cite{hemachandra2023trainingfreeneuralactivelearning} in order to optimize an LLM/MLLM's performance metric $C$. 
By doing so, deeper insights into the underlying scaling laws governing how model architecture and data selection jointly influence the LLM/MLLM's performance metric $C$ can be developed.

    \section{Inference Optimization}
    \label{sec:inference}

Optimizations carried out at the inference stage significantly affect the performance of LLMs.
For example, given a trained LLM, it is common practice to provide a prompt (i.e., a snippet of text) that the LLM uses to generate further text conditioned on the snippet.
This represents a forward process $\T{\F{\Train}, \Inference} \rightarrow C$ in~\cref{eq:T}, where the prompt is a component of inference ingredient 
$\mathcal{I}$, and inverting the process to carefully construct prompts that can instruct the LLM to perform a specific downstream task, hence achieving a desired performance measured by the metric $C$,  is challenging.
Thus, inference optimization can be viewed as an inverse problem of $\T{\F{\Train}, \Inference} \rightarrow C$ in~\cref{eq:T}, where the goal is to design inference schemes in $\Inference$ that, when combined with a model trained on $\Train$, achieves the desired performance metric $C$.
Furthermore, one can also aim to uncover the underlying scaling laws at inference time with respect to optimized data, model architecture, and computational resources.

\subsection{Data Optimization at Inference Time}
Prompts are key components of $\mathcal{I}$ during the LLM inference.
A widely adopted popular prompting structure consists of instructions and few-shot demonstrations (data samples), also known as exemplars.
This approach leverages LLMs' ability for \textit{in-context learning}, which has emerged with the rapid scaling of LLMs in terms of the number of parameters, particularly since the advent of GPT-3~\citep{brown2020fewshot}.
Specifically, the LLMs can understand and perform tasks based on exemplars and instructions provided only in the context of the prompt, without relying on conventional training methods like fine-tuning on specific datasets~\citep{liu2022makes}. 
It is widely observed that the design of instructions and the selection of exemplars in the prompt significantly influence the LLM performance~\citep{rubin2022retrieval,TMLR24_albalaksurvey,wu2024ease}.

Prompting techniques have been introduced to steer the LLM responses towards better accuracy, tailored tone, improved focus~\citep{Anthropic-system-prompts,Anthropic-role-of-system-prompts}, and reduced hallucinations~\citep{apple-system-prompt,xu2024hallucination}.
In short, prompting is a tool to achieve the desired performance metric $C$. 
Despite its benefits, designing instructions and selecting exemplars for prompts typically requires a human-intensive and costly trial-and-error approach~\citep{mishra2021reframing,reynolds2021manual}.
Recent works have explored heuristic local search methods~\citep{ICLR23_zhou2022large} and evolutionary strategies~\citep{ACL23_prasad2023grips,guo2023connecting} to identify the best instructions and retrieval-based methods to find the most relevant exemplars~\citep{liu2022makes,rubin2022retrieval}.
However, these methods can still be costly and sub-optimal, raising the important question: How can prompts be efficiently optimized under resource constraints, such as limited computational resources or fewer queries?

Viewing the research question as an inverse problem, one can formulate the prompt optimization problem 
as a black-box optimization problem where the inputs are the prompts (comprising instructions and exemplars) and the output is the prompt's performance.
Then, optimization techniques such as the NeuralUCB algorithm can be applied to optimize the prompt for the best performance under resource constraints~\citep{zhou2020neural,dai2022sample}.
Specifically, in the NeuralUCB algorithm, 
a neural network is trained on past observations to predict the LLM performance for different combinations of instructions and exemplars.
This approach will help uncover underlying scaling laws and understand the effect of instructions and exemplars on LLM performance.
Moreover, finding the exemplars (given a fixed budget) and instructions to achieve the best LLM performance helps to uncover the scaling law of LLM performance with respect to the number of exemplars used in $\Inference$. 
This scaling law will allow the real applications to choose the minimal number of exemplars to achieve a target performance metric $C$.

Since both the data in $\Train$ and the data in $\Inference$ affect the final LLM performance, optimizing an LLM's performance requires the \textit{joint optimization} of in-context data in $\Inference$ and training data in $\Train$. 
To efficiently solve this optimization problem, we further advocate for research into developing algorithms that automatically select the optimal combination of in-context and training data for an LLM. This approach will help us to uncover fundamental scaling laws governing the combined impact of both training $\Train$'s and inference $\Inference$'s ingredients on the performance metric $C$.

Additionally, we can consider the problem of prompt optimization with human feedback, aiming to minimize the amount of human feedback required to find the best prompt that maximizes LLM performance.
Specifically, we consider the inverse problem in which the performance metric $C$ is defined as the alignment of LLM responses with human values, such as helpfulness. The goal is to optimize the prompt to improve the alignment.
Recent works have shown that humans are better at providing preference feedback than giving a score, which has been the focus of prior prompt optimization works~\citep{lin2024instinct,hu2024zerothorder,wu2024ease,zhou2024detail}. To address this, recent works propose a framework of prompt optimization that relies solely on human preference feedback on the LLM responses~\citep{lin2024promptopt}, demonstrating superior performance compared to prior results on prompt optimization.

\subsection{Model Optimization at Inference Time} 
When deploying resource-efficient LLMs, understanding the scaling laws for determining optimal model configurations is crucial for effective and efficient usage~\cite{devvrit2024matformer}. Selecting the best model configuration during inference is a critical inverse problem that aims to identify an LLM setup capable of achieving a target performance metric $C$ with minimal computational resources.
Formally, the inference-time model configuration should be considered as part of the inference ingredients $\Inference$ in~\cref{eq:T}. The goal is to identify a model configuration that minimizes computational requirements while achieving the desired performance metric $C$. 
As model sizes increase, they require proportionately more compute and memory per generation, making them impractical in resource-constrained settings. 
Furthermore, simply scaling model parameters does not guarantee better performance, especially in scenarios constrained by the variety and quality of available data~\citep{allen2020towards}.

This challenge can be addressed from two perspectives: (1) selecting the optimal model at inference time from LLMs of varying sizes and capacities using methods like model valuation and selection \cite{xu2024model}, and (2) determining the optimal number of activated routes in Mixture-of-Experts (MoE) LLMs during inference to balance efficiency and performance.
By systematically exploring model size scaling, one can determine how to adjust the model size to meet the demands of specific tasks during inference.
Ultimately, uncovering the scaling laws for model scaling at inference allows trade off between computational efficiency and performance.

\subsection{Compute Optimization at Inference Time}
The introduction of OpenAI's o1 model and DeepSeek R1, which are designed to facilitate CoT reasoning during inference, has induced increasing interest in scaling computational resources at inference to improve model performance~\citep{snell2024scalingllmtesttimecompute,wu2024inferencescalinglawsempirical,zhou2025mem1}. 
Existing work~\citep{chen2024simple} has demonstrated a scaling law that relates model performance to the computational resources used during inference. 
But it focuses only on a single inference scheme, where the inference scheme (e.g., CoT) is an inference ingredient $\Inference$ in~\cref{eq:T}.
Besides CoT, other inference schemes, such as prompt optimization, optimization with human feedback, retrieval-augmented generation~\citep{gao2024retrievalaugmentedgenerationlargelanguage, shao2024scalingretrievalbasedlanguagemodels,shao2025reasonir}, repeated sampling~\citep{brown2024largelanguagemonkeysscaling, gui2024bonbonalignmentlargelanguage}, and ensemble models~\citep{allen2020towards}, have also been explored to scale inference-time compute for improving LLM performance. 

An exciting area of research is to optimize a mix of these inference schemes within a fixed computational budget, uncovering more effective model scaling behavior. 
Specifically, computational resources can be quantified by the number of responses generated by each of these inference schemes. 
By optimally allocating resources across schemes and then strategically selecting and merging these responses improves LLM performance. 
Studying how the scaling law changes when inference schemes are optimally combined will provide deeper insight into the computational requirements necessary to achieve a target performance $C$.

Apart from performance optimization, another area of research where inverse problem formulation can be utilized is inference time optimization. Existing research~\citep{leviathan2023,spector2023stagedSD,wu-etal-2025-tetris} has leveraged inverse methods to identify the optimal configuration of speculative decoding. These applications demonstrate the potential of applying scaling laws to improve model inference speed.

\subsection{Joint Optimization at Inference Time} 
LLM performance is influenced by a complex interplay between data, model, and compute. Given a fixed computational cost specified by the performance metric $C$, it is crucial to identify the optimal combination of model configuration and inference schemes when user prompts (i.e., data) are fixed. Thus, jointly optimizing the model configuration and inference schemes can help to approach optimal LLM performance. 
Specifically, exploring how to allocate computational resources across different inference schemes and models should be a key focus.
This approach will help uncover the underlying scaling laws that characterize how models, inference schemes, and computational budgets collectively impact LLM performance.
These scaling laws can help to decide minimal model parameters and computational resources needed for LLMs to achieve desired performance, reducing the serving cost of these models in real-life applications.

    \section{Unlearning}
    \label{sec:unlearning}

Machine unlearning (MU) is the process of removing the influence of a set of training data (i.e.,  erased data) from a trained model to either comply with data owners' deletion requests~\citep{gdpr16, ccpa18} or erase harmful data to improve the model performance~\citep{fore2024unlearning, liu2024towards, zhou2024making}.
We consider two inverse problems.
\textbf{Verification of MU} is an inverse problem of $\F{\Train} \rightarrow \LLM$ as given any ``unlearned'' model, aiming to identify if the erased data is present in the training ingredients $\Train$.
\textbf{MU techniques} can also be viewed as an inverse problem of $\T{\F{\Train}, \Inference} \rightarrow C$. 
Given certain performance metrics (e.g., poor knowledge on weapons of mass destruction \cite{li2024wmdp}, similar performance on the retained data as before unlearning), the goal is to design the inference ingredients (e.g., unlearning prompts) in $\Inference$, or to adjust the datasets and training procedure (e.g., use of training checkpoints, model architecture that facilitates unlearning without retraining) in $\Train$ to achieve the desired performance metrics.

\subsection{MU Verification}
Despite the growing interest in MU for LLMs~\citep{eldan2023s, chen-yang-2023-unlearn, liu2024rethinking}, one major challenge remains: How to efficiently verify whether the requested data is not present in an unlearned LLM?
At first glance, we can compare the similarity of an unlearned LLM with the model trained only on the retained data (without the erased data) \cite{nguyen2022survey, maini2024tofutaskfictitiousunlearning}. However, such an approach requires obtaining the LLMs retrained only on the retained data, which is computationally expensive~\citep{yao2024machine} or infeasible when there are computational hardware constraints.
Other MU metrics try to address the challenge empirically. For example, the Membership Inference Attack (MIA) metric~\citep{shokri2017membership} expects low accuracy on the erased data when assessed by an adversarial model trained to classify whether data points were members of the training dataset. These metrics fall short as they either require white-box access to the LLM~\citep{duan2024membership}, which is often unfeasible, or require training shadow models, which are computationally expensive~\citep{shokri2017membership}. Furthermore, the MIA metric depends on the adversarial model's ability to distinguish between membership and non-membership~\citep{duan2024membership}, which can be limited when similar data points are present in both erased and retained data (e.g., multiple news sources reporting on the same event).
Thus, such a situation raises the following question: How can an efficient MU verification metric for LLMs not requiring model retraining be designed? Can the metric be intuitive and effective despite the presence of similar data?

\begin{figure}[!ht]
    \centering
    \includegraphics[width=\columnwidth]{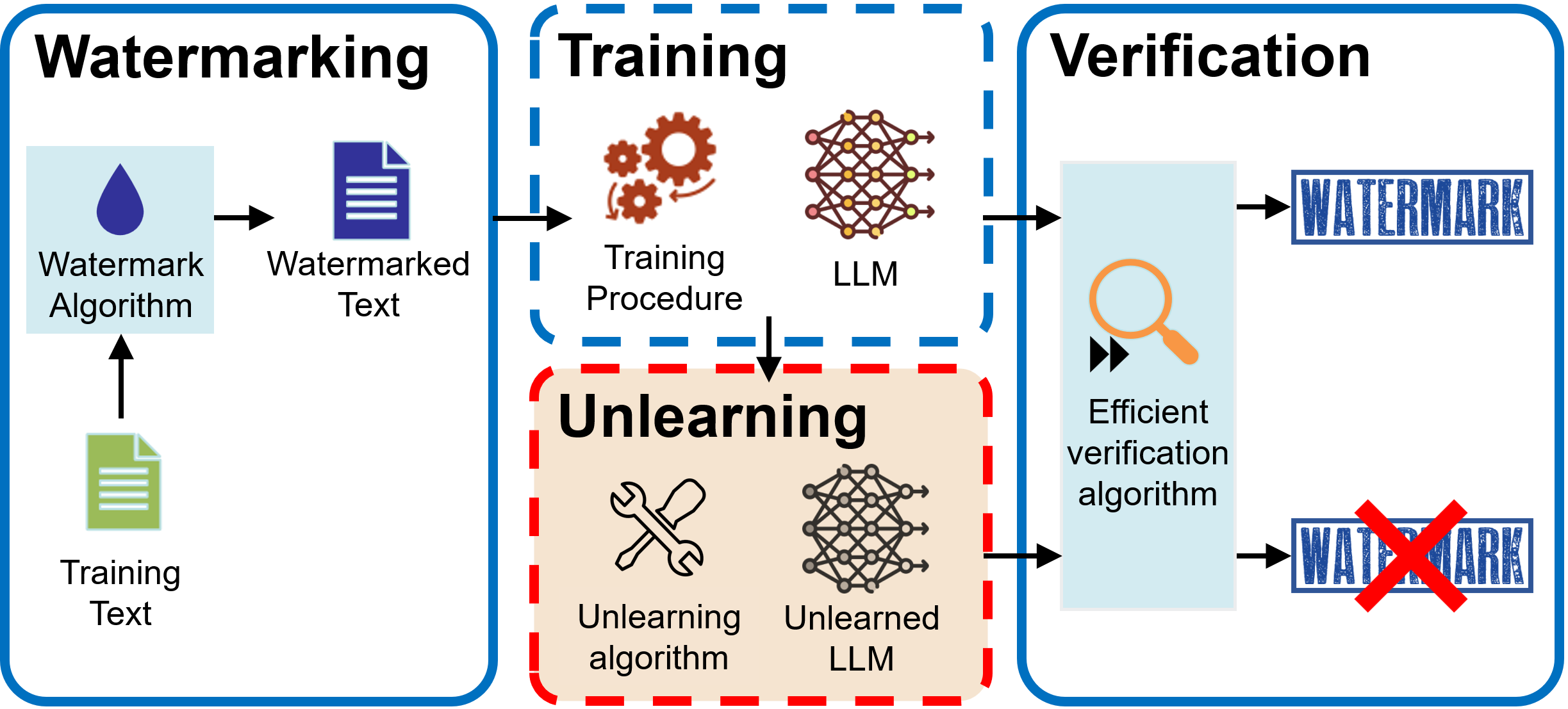}
    \caption{Watermarking as an unlearning metric.}
    \label{fig:unlearning} 
\end{figure}

Answering these open questions is non-trivial.
One potential approach is to leverage related work on scalable and robust watermarking~\cite{lau2024waterfall} for text data, by embedding unique watermarks into each data owner's text content before LLM training \citep{lu2025waterdrum}. 
Such watermarks should remain detectable and verifiable in LLM predictions after fine-tuning and, hence, be used to test the effectiveness of unlearning.
An overall evaluation framework is illustrated in Fig.~\ref{fig:unlearning}.
Research on this metric could help support the scaling law that retraining-free metrics require data attribution to trace the impact of individual data points during initial training, thereby \emph{improving unlearning procedures without the need for complete retraining}.

MU metrics can help define scaling laws governing the difficulty of unlearning erased data. 
Previous work \cite{unlearning_hard} explored how the tug-of-war (ToW) verification metric, which compares the accuracies of the unlearned and retrained models, is influenced by the properties of erased and retained data. It also examined how certain properties of erased data, like high memorization score, may require different MU techniques to achieve a better ToW score.
Building on these works, one can further explore how this new retraining-free metric and other MU metrics are influenced by dataset properties, such as size, watermark count, and the similarity between erased and retained data. 
These insights will uncover underlying scaling law that guides the selection of MU techniques and improve the reliability of metrics used for evaluating unlearning techniques.

\subsection{MU Techniques} 
Many existing MU techniques modify model weights~\citep{chen-yang-2023-unlearn, yao2023large, jang2023knowledge}, making them unsuitable for black-box LLMs or when fine-tuning is expensive due to computational constraints. 
While recent approaches such as offset unlearning~\citep{huang2024offset} can be applied to black-box models but often cause an unacceptable performance drop on the retained data~\citep{huang2024offset}.
Prior LLM work on in-context unlearning~\cite{pawelczyk2023context} is further limited to sentiment classification and does not scale to generative tasks.
Existing MU techniques may perform well on metrics like MIA but risk unlearning some retained data that are similar to the erased data~\citep{jin2024rwku}.
This raises a critical question: Is post-hoc unlearning (i.e., only modifying $\Inference$) feasible for text generation without compromising the performance of the retained data or introducing unintended biases?

The target performance $C$ of an LLM is defined as minimizing the generation of harmful data or weak watermark strength, measured by the watermarking-based MU metric while retaining its performance on other metrics, such as the validation loss. 
A key question is how to achieve $C$ efficiently by modifying the inference process $\Inference$.
We advocate for research that identifies the private or harmful data (e.g., by detecting the watermarks present in generated text) and adaptively modifies $\Inference$ during inference to suppress their influence and prevent such data from being generated.

Alternatively, can $C$ be achieved efficiently by modifying the model architecture in $\Train$ such that it is easier to unlearn? One possible approach is using the intrinsic sparsity of MoE transformer paradigm \cite{shazeer2017outrageously, lepikhin2020gshard, fedus2022switch} to isolate the influence of data to a few experts and thereby perform unlearning more efficiently on fewer model parameters.
Overall, the goal is to improve LLM performance on the given metric $C$ and uncover underlying scaling laws for unlearning during inference.
Specifically, this involves identifying how the metric $C$, such as the loss on the erased and retained data, varies with the size of these datasets, computation cost, and model's ability to unlearn during inference.
These scaling laws can help identify the most suitable MU techniques for removing harmful knowledge from LLMs and determine how much data can be erased before performance metrics drop below a predefined threshold that necessitates retraining.

    \section{Conclusion and Future Outlook}
    \label{sec:summary}

\vspace{-0.75mm}
This position paper highlights the importance of understanding of the scaling laws that govern the behavior of LLMs, such as data requirements and compute scaling laws.
To uncover the underlying scaling laws, we advocate for research exploring two classes of inverse problems for LLMs (i.e.,~\cref{eq:F} and~\cref{eq:T}): Identifying optimal input ingredients and achieving desired performance metrics by adjusting both training and inference ingredients.
Specifically, we frame data selection, inference optimization, and machine unlearning as inverse problems, each presenting unique challenges to solve.
Yet, jointly optimizing them (including data, model architecture, training procedures, inference scheme, and unlearning techniques) holds great potential for advancing the development and deployment of LLMs.

Instead of iterating over the engineering efforts to further improve the empirical performance, we advocate to uncover the underlying fundamental scaling laws governing the training and inference of LLMs via inverse problems, which can lay the foundations for building better LLMs. 
These scaling laws can improve specific applications by providing better selection methods for training data and model architectures, flexible unlearning techniques, methods with improved inference efficiency, and optimized inference schemes. 

Looking ahead, future research should explore these scaling laws and investigate how the interplay among various components and ingredients impacts overall performance. 
Emerging technologies and methodologies from fields like optimization theory can provide novel tools for tackling inverse problems in LLMs. 
Additionally, advancements in machine unlearning will be crucial as models become more complex, ensuring they can adapt without compromising functionality or privacy standards.
By integrating these approaches, we may uncover innovative solutions that improve the efficiency and cost-effectiveness of LLM development.

    \section*{Limitations}
    While the inverse problem formulation offers a promising perspective for studying large language models (LLMs), it is important to recognize that not all problems in LLMs have lend themselves to well-defined inverse formulations. Analogous to how the inverse for a many-to-one function is ill-defined mathematically, many forward problems in LLMs, such as data aggregation or input-to-output mappings, are inherently many-to-one. This leads to potential ambiguity or ill-posedness in their inverse counterparts. Addressing these challenges will require further theoretical and methodological advancements. 
    
    Additionally, this paper focuses on a limited set of illustrative problems, such as data selection, inference optimization, and machine unlearning for LLMs, to demonstrate the potential of the inverse problem framework. A comprehensive exploration of its applicability across the broader and rapidly evolving landscape of LLM research remains an open direction. We encourage future work to uncover additional problem domains where inverse formulations may offer meaningful insights.

    \section*{Ethic Statement}
    LLMs are largely trained on data scraped from the Internet, which may include dangerous, unsafe, biased, or inaccurate content. As a result, LLMs risk reproducing these harmful patterns in their generated outputs. Moreover, the use of scraped data raises both legal and ethical issues. The data may be copyrighted or include sensitive personal information without the consent of the data subjects. In response, we aim to mitigate these risks by improving data selection and developing machine unlearning techniques that support the removal of harmful or sensitive data and machine unlearning verification metrics to verify removal.

    \section*{Acknowledgments}
    This research/project is supported by the National Research Foundation, Singapore under its National Large Language Models Funding Initiative (AISG Award No: AISG-NMLP-2024-001). Any opinions, findings and conclusions or recommendations expressed in this material are those of the author(s) and do not reflect the views of National Research Foundation, Singapore.

    \bibliography{references}

\end{document}